\documentclass[letterpaper, 10 pt, conference]{ieeeconf}  

\IEEEoverridecommandlockouts                              

\usepackage{algorithm}
\usepackage{algpseudocode}
\usepackage{amssymb}
\usepackage{amsmath}
\usepackage{bm}
\usepackage{mathtools}
\usepackage{pgfplots}
\usepackage{pgfplotstable}
\usepackage{booktabs}
\pgfplotsset{compat=1.18}
\usepgfplotslibrary{groupplots}
\usepackage{cite}
\usepackage{graphicx}
\usepackage{subcaption}
\usepackage{adjustbox}
\pgfplotsset{compat=1.18}

\DeclareMathOperator*{\argmax}{arg\,max}

\pgfplotsset{
  bwplot/.style={
    width=0.36\textwidth,
    height=4.6cm,
    ytick={0,0.2,0.4,0.6,0.8,1.0},
    enlarge x limits=0.05,
    grid=major,
    grid style={dashed, gray!30},
    tick label style={font=\footnotesize},
    label style={font=\small},
    title style={font=\small, yshift=-1pt},
    legend style={
      font=\footnotesize,
      draw=none, fill=none,
      inner sep=1pt,
      /tikz/every even column/.append style={column sep=6pt},
    },
    legend cell align=left,
    error bars/error bar style={black, solid, line width=0.4pt},
    error bars/error mark options={
      rotate=90, mark size=2pt, line width=0.4pt, black, solid,
    },
    cycle list={
      {black, solid,    mark=*,         mark size=1.2pt, mark options={solid}},
      {black, solid,    mark=square*,   mark size=1.2pt, mark options={solid}},
      {black, dashed,   mark=triangle*, mark size=1.6pt, mark options={solid}},
      {black, dashed,   mark=diamond*,  mark size=1.6pt, mark options={solid}},
      {black, dotted,   mark=otimes*,   mark size=1.6pt, mark options={solid}},
    },
  },
}

\title{\LARGE \bf
An Evidence Hierarchy for Bayesian Object Classification via OSINT-Aided Heterogeneous Sensor Fusion
}

\author{Jan Nausner$^{1}$ and Michael Hubner$^{1}$
\thanks{$^{1}$Center for Digital Safety \& Security, Austrian Institute of Technology GmbH (AIT), Giefinggasse 4, 1210 Vienna, Austria {\tt\small \{firstname\}.\{lastname\}@ait.ac.at}}%
\thanks{This work was funded by the European Union under the European Defence Fund grant 101121233 (DeterMine).}
}

\begin{document}

\maketitle
\thispagestyle{empty}
\pagestyle{empty}
\begingroup
\renewcommand\thefootnote{}
\footnotetext{%
© 2026 IEEE. Accepted for the 2026 IEEE International Conference on Multisensor Fusion and Integration (MFI 2026).
}
\endgroup

\begin{abstract}

Heterogeneous sensor fusion is vital for detecting, localizing, and classifying CBRNE threats. However, individual sensors are often only capable of detecting a subset of relevant threats with varying reliability or can even provide only indirect threat indications, making threat classification challenging. Furthermore, high clutter rates on the sensor side present a great challenge for fusion systems. Additionally, the limited availability of high quality datasets hinders the advancement of learning-based detection and classification models in smart sensors. To mitigate these sensor related shortcomings, a context-aware and domain knowledge-enhanced fusion process is proposed. First, a novel evidence hierarchy is established that enables modeling of direct, indicative, and contextual information. Second, contextual information about the environment is introduced into the fusion process, by collecting, processing, and exploiting OSINT inputs. Third, all levels of the evidence hierarchy are used to craft a Bayesian threat type classification mechanism with domain knowledge-informed priors. The proposed methodology is evaluated in simulated scenarios, and the results demonstrate the benefit of the proposed fusion approach in terms of robustness to clutter and prior mismatch, with an overall classification accuracy of up to 95\%.

\end{abstract}

\section{INTRODUCTION}\label{sec:introduction}

The detection, localization, and classification of chemical, biological, radiological, nuclear, and explosive (CBRNE) threats represent very demanding tasks for multi-sensor fusion systems. Operational environments impose numerous challenges such as: high clutter rates may generate false alarms, individual sensors covering only subsets of relevant threat types with varying reliability, and some sensors providing only indirect anomaly-based indications rather than direct type classification. These challenges are further aggravated by the scarcity of high-quality labeled datasets, which limits the applicability of purely learning-based approaches.

Multi-sensor fusion has long been recognized as the main strategy to overcome individual sensor limitations \cite{hall2002introduction, luo2002multisensor}. However, existing fusion frameworks do not explicitly distinguish and leverage the different levels of evidence — direct type distinction, indirect anomaly-based indication, or environmental context provision. Similarly, while contextual information has been identified as beneficial for improving fusion performance \cite{snidaro2015context}, its systematic integration into CBRNE classification pipelines remains limited.

\subsection{Contributions}

In this work, a context-aware, domain knowledge-enhanced heterogeneous Bayesian sensor fusion method for CBRNE threat type classification is proposed. The main contributions are: (i) a formal evidence hierarchy that explicitly separates direct, indicative, and contextual evidence; (ii) collection and processing of relevant open-source intelligence (OSINT) from geospatial big data (GBD) and exploiting it as contextual evidence; and (iii) a Bayesian maximum a posteriori (MAP) classifier that jointly exploits all evidence levels within a unified probabilistic model.

\subsection{Related Work}

Various models \cite{hall2002introduction, luo2002multisensor} establish a fundamental multi-level view of data fusion, motivating a hierarchical treatment of evidence. Although context-based fusion \cite{snidaro2015context} is well-recognized as an enabler — demonstrated concretely through tracking refinement with environmental data \cite{visentini2011integration} and terrain-conditioned sensor detections \cite{riese2018fusion, sheppard2024learning} — its application to the CBRNE use-case remains limited. The proposed evidence hierarchy addresses this by formally separating direct, indicative, and contextual evidence within a single framework.


GBD sources were shown to provide effective spatial evidence for scene classification tasks ranging from urban functional zone mapping \cite{lu2022unified} to population estimation \cite{cheng2021remote} and land use classification \cite{srivastava2019understanding}, confirming GBD as a mature complementary information source \cite{li2022deep}. In this work, this paradigm is transferred to the CBRNE domain, using GBD-derived regions as the geospatial OSINT basis for domain knowledge-informed threat type priors.

Decision-level fusion methods — including Bayesian inference, Dempster-Shafer evidence theory, and voting — have been extensively compared for threat/clutter discrimination \cite{frigui2012evaluation} and type classification \cite{stanley2002feature, prado2015bayesian}, establishing sensor confidence as a highly important parameter for the fusion process. Learning-based approaches \cite{qiu2023joint, jafuno2025classification} push differentiation performance, but are fundamentally constrained by scarce labeled CBRNE data \cite{malizia2025mineinsight}, motivating the model-based, domain knowledge-driven strategy adopted in this manuscript. CBRNE multi-sensor fusion of UAV-borne gamma measurements and LiDAR terrain data is demonstrated in \cite{schraml2022real}, illustrating the benefit of contextual evidence sensors in the proposed hierarchy.

\section{METHODOLOGY}\label{sec:methodology}

\subsection{Evidence Hierarchy}\label{sec:evidence-hierarchy}

To categorize various levels of information, either from heterogeneous sensors or based on contextual cues, an evidence hierarchy is introduced. This enables the formal modeling of the various evidence levels in order to integrate them in a context-aware multi-sensor fusion process for object classification. The evidence hierarchy entails three levels: direct, indicative, and contextual.

\subsubsection{Direct Evidence} Provided by measuring intrinsic object characteristics to enable direct object type discrimination, hence detections include the predicted types.
\subsubsection{Indicative Evidence} Generated by measuring correlational evidence such as cause and effect or anomalies. Detections usually only indicate the presence of an object, and, instead of direct type discrimination, only a set of possible object types is provided (explicitly or implicitly).
\subsubsection{Contextual Evidence} Comprises evidence that is not directly related to the object but rather accounts for the object environment \cite{snidaro2015context}. Provides contextual cues acting supplementary to the other evidence levels and in this way can aid in object detection and classification. Some examples:
(i) Detection reliability: A sensor detection may be more or less reliable, or the sensor could even be not functional at all, given the region in which the sensor was operated.
(ii) Type prevalence: Objects of a specific type can be more or less common given the region in which a detection was generated.

\subsection{Problem Statement}\label{sec:problem-statement}

\begin{table*}[t]
\caption{Mapping of information sources to the evidence hierarchy.}
\label{tab:evidence-hierarchy}
\centering
\footnotesize
\begin{tabular}{l|l|l}
\textbf{Evidence level} & \textbf{Source} & \textbf{Information content} \\
\hline
Direct & Sensor with object type prediction & $P(D,t|Z),\ P(D|t)$ \\
Indicative & Sensor without object type prediction & $P(D|Z),\ P(D|t)$ \\
Contextual & OSINT-derived region with domain knowledge-informed prior & $P(t|r)$ \\
\end{tabular}%
\end{table*}

A set of static CBRNE threat objects $k \in \mathcal{K} = \{1,\dots,N_K\}$, with types $t^{(k)} \in \mathcal{T} = \{T_1,\dots,T_{N_T}\}$ and unknown positions $\mathbf{x}^{(k)} = [x^{(k)}, y^{(k)}]^T$, is distributed over an arbitrarily shaped region of interest (ROI). The ROI is subdivided into non-overlapping regions with types $r \in \mathcal{R} = \{R_1,\hdots,R_{N_R}\}$. The regions are defined by the process described in Sec. \ref{sec:contextual-evidence}. For each region type $r$ and each threat type $t$ the prior probability $P(t|r)$ is defined, indicating type prevalence per region type.

A set of sensors $s \in \mathcal{S} = \{S_1,\dots,S_{N_S}\}$, stationary or mobile, scan the entire ROI. Each sensor is only capable of detecting a subset of threat types $\mathcal{T}^{(s)} \subseteq \mathcal{T}$ with type-dependent detection probabilities $P_D^{(s,t)}=P^{(s)}(D=1|t), t\in\mathcal{T}^{(s)}$ ($D=1$ denotes the hypothesis that the sensor return results from a true object, while $D=0$ denotes a clutter return). These probabilities can be set empirically or based on domain knowledge. Furthermore, the set of sensors can be subdivided into two categories $\mathcal{S} = \mathcal{S}_D \cup \mathcal{S}_I$, where $\mathcal{S}_D$ includes sensors that provide direct evidence and $\mathcal{S}_I$ includes sensors that provide indicative evidence, as defined in Sec. \ref{sec:evidence-hierarchy}.

The detection of object $k$ generated by a direct evidence sensor $s \in \mathcal{S}_D$ is characterized by the tuple $Z^{(k,s)} = (\pi_n^{(k,s)}, \hat{t}^{(k,s)})$,
whereas the return of the same object generated by an indicative evidence sensor $s \in \mathcal{S}_I$ is given by $Z^{(k,s)} = \pi_n^{(k,s)}$.
The label $\hat{t}^{(k,s)}$ is the type of object predicted by a sensor $s\in\mathcal{S}_D$. For sensors $s\in\mathcal{S}_D$ the confidence
\begin{equation}
    P(D=1,t|Z^{(k,s)}) = \begin{cases}
        \pi^{(s)},\ &t=\hat{t}^{(k,s)}\\
        0,\ &t \neq \hat{t}^{(k,s)},
    \end{cases}
\end{equation}
denotes the probability of a true detection with a specific type. For sensors $s\in\mathcal{S}_I$ the confidence $P(D=1|Z^{(k,s)}) = \pi_n^{(k,s)}$ represents the probability of a true detection. Typically, $\pi_n^{(k,s)}$ relates to sensor-intrinsic characteristics, such as the signal to noise ratio (SNR) or classifier properties. Tab. \ref{tab:evidence-hierarchy} summarizes the categorization of the information sources according to the evidence hierarchy proposed in Sec. \ref{sec:evidence-hierarchy}.

In this work, it is assumed that every sensor creates at most one return per object and that all sensor detections resulting from the same object $k$ are grouped by a data association method (such as described in \cite{nausner2026soda}):
\begin{equation}
\begin{split}
    \mathcal{Z}^{(k)} = \{Z^{(k,s_1)},Z^{(k,s_2)},\hdots\},\\
    s_1,s_2,\hdots \in, \mathcal{S},\ s_1\neq s_2,\hdots.
\end{split}
\end{equation}
Additionally, the region of the object $r^{(k)}$ is provided based on the estimated object position $\hat{\mathbf{x}}^{(k)}$ returned by the data association step. The goal is to predict the threat type $\hat{t}^{(k)} \in \mathcal{T}$ for each object $k$ based on the object region $r^{(k)}$ and the available sensor detections $\mathcal{Z}^{(k)}$, comprising direct and/or indicative evidence.

\subsection{Object Type Classification}\label{sec:type-classification}

To perform threat object type classification, this work relies on Bayesian inference \cite{blackman1999design, bishop2006pattern}. Note that in the following derivations, the threat object index $k$ will be dropped in favor of compactness and readability.

\subsubsection{Threat Type Posterior}
Given the set of detections $\mathcal{Z}$ of an object (referred to as $\mathcal{Z}^{(k)}$ in Sec. \ref{sec:problem-statement}), resulting from a data association process (such as \cite{nausner2026soda}), and the regional prior $P(t|r)$, determined by the region type $r\in\mathcal{R}$ of the region holding the object, the threat type posterior of the object is given by 
\begin{equation}\label{eq:compact-posterior}
    P(t|\mathcal{Z},r) = \frac{P(t|r)P(\mathcal{Z}|t,r)}{\sum_{t\in\mathcal{T}} P(t|r)P(\mathcal{Z}|t,r)}.
\end{equation}
Assuming conditional independence of all detections $Z^{(s)}\in\mathcal{Z}$ given type $t$ and that the regions are non-informative for the detections, the likelihood can be factorized over all contributing sensors: $P(\mathcal{Z}|t,r) = \prod_{s:Z^{(s)}\in\mathcal{Z}}P(Z^{(s)}|t)$.
Applying the law of total probability to the variable $D\in\{0,1\}$ expands the per-sensor likelihood:
\begin{equation}
\begin{adjustbox}{max width=\columnwidth}
$
    P(Z^{(s)}|t) = \Lambda(Z^{(s)},t) = \sum\limits_{d\in\{0,1\}}P(Z^{(s)}|D=d,t)P(D=d|t).
$
\end{adjustbox}
\end{equation}
The type-dependent sensor prior is found in $P(D=1|t)=P_D^{(s,t)}$ and $P(D=0|t)=1-P_D^{(s,t)}$. To further develop the per-sensor likelihood, a case distinction between direct and indicative evidence must be made:
\begin{equation}
    P(\mathcal{Z}|t,r) = \prod\limits_{s_D} \Lambda_{s_D}(Z^{(s_D)},t) \prod\limits_{s_I} \Lambda_{s_I}(Z^{(s_I)},t),
\end{equation}
where $s_D \vcentcolon = s:s\in\mathcal{S}_D\wedge Z^{(s)}\in\mathcal{Z}$ and $s_I \vcentcolon= s:s\in\mathcal{S}_I\wedge Z^{(s)}\in\mathcal{Z}$.

\subsubsection{Direct Evidence Sensor Likelihood}
If $s\in\mathcal{S}_D$, then $Z^{(s)}$ also includes a predicted object type $\hat{t}^{(s)}$. By Bayes theorem and assuming a uniform sensor-internal prior $P(Z^{(s)})$:
\begin{equation}
    P(Z^{(s)}|D=d,t)\propto\frac{P(D=d,t|Z^{(s)})}{P(D=d,t)}.
\end{equation}
Let $d=1$. In Sec. \ref{sec:problem-statement} it is stated that $P(D=1,t|Z^{(s)}) = \pi^{(s)}$, therefore,
\begin{equation}
    P(Z^{(s)}|D=1,t) = \begin{cases}
        \frac{\pi^{(s)}}{P(D=1,\hat{t}^{(s)})} \propto \pi^{(s)},\ &t=\hat{t}^{(s)}\\
        0,\ &t \neq \hat{t}^{(s)},
    \end{cases}
\end{equation}
as the denominator is now a constant.
Let $d=0$. Since clutter generation is independent of the object type, 
$P(Z^{(s)}|D=0,t) = P(Z^{(s)}|D=0) \propto P(D=0|Z^{(s)})$ by Bayes theorem with assumed uniform sensor-internal prior. Marginalization of $t$ yields
\begin{equation}
\begin{split}
    P(D=0|Z^{(s)}) &= 1 - P(D=1|Z^{(s)})\\
    & = 1-\sum_{t'} P(D=1,t|Z^{(s)}) = 1 - \pi^{(s)}.
\end{split}
\end{equation}
The likelihood is hence given by
\begin{equation}
\begin{adjustbox}{max width=\columnwidth}
$
    \Lambda_{s_D}(Z^{(s)},t) =
    \begin{cases}
        \pi^{(s)}P_D^{(s,t)}+(1-\pi^{(s)})(1-P_D^{(s,t)}),\ &t=\hat{t}^{(s)}\\
        (1-\pi^{(s)})(1-P_D^{(s,t)}),\ &t\neq\hat{t}^{(s)}.
    \end{cases}
$
\end{adjustbox}
\end{equation}

\subsubsection{Indicative Evidence Sensor Likelihood}
Given that for $s\in\mathcal{S}_I$ the sensor detection $Z^{(s)}$ does not provide any type discriminatory information, it holds that $P(Z^{(s)}|D=d,t)=P(Z^{(s)}|D=d)$. Again, by Bayesian inversion and assuming a uniform sensor-internal prior $P(Z^{(s)})$, set
\begin{equation}
    P(Z^{(s)}|D=1)\propto P(D=1|Z^{(s)})=\pi^{(s)}
\end{equation}
and
\begin{equation}
\begin{adjustbox}{max width=\columnwidth}
$
    P(Z^{(s)}|D=0)\propto P(D=0|Z^{(s)})=1-\pi^{(s)}.
$
\end{adjustbox}
\end{equation}
This yields the likelihood
\begin{equation}
\begin{adjustbox}{max width=\columnwidth}
$
    \Lambda_{s_I}(Z^{(s)},t) = \pi^{(s)}P_D^{(s,t)}+(1-\pi^{(s)})(1-P_D^{(s,t)}).
$
\end{adjustbox}
\end{equation}

\subsubsection{MAP Threat Classification}
The complete threat type posterior, developed from Eq. \ref{eq:compact-posterior}, is hence given by
\begin{equation}
\begin{adjustbox}{max width=\columnwidth}
$
    P(t|\mathcal{Z},r) = \frac{P(t|r)\prod_{s_D} \Lambda_{s_D}(Z^{(s_D)},t) \prod_{s_I} \Lambda_{s_I}(Z^{(s_I)},t)}{\sum_{t\in\mathcal{T}} P(t|r)\prod_{s_D} \Lambda_{s_D}(Z^{(s_D)},t) \prod_{s_I} \Lambda_{s_I}(Z^{(s_I)},t)}.
$
\end{adjustbox}
\end{equation}
Finally, the predicted threat type is determined by the maximum a posteriori estimate over:
\begin{equation}
    \hat{t} = \argmax_{t' \in \mathcal{T}}P(t=t'|\mathcal{Z},r).
\end{equation}

\subsubsection{Computational Complexity}
The computational complexity of classifying $N_K$ threats with the method presented above is in $\mathcal{O}(N_K \cdot N_T \cdot N_S)$. Since the number of threat types and the number of sensors is usually constant during a scan, the overall computational complexity of classifying all threats is linear in the number of threat objects $N_K$.

\begin{table}[h]
\caption{Basic scenario parameters.}
\label{tab:basic_params}
\centering
\resizebox{\columnwidth}{!}{%
\begin{tabular}{l|ccc|l}
\hline
\textbf{Sensor prior $P_D^{(s,t)}$} & $t=A$ & $t=B$ & $t=C$ & Evidence level \\
\hline
$s=S_1$ & 0.9 & 0.4 & 0 & indicative \\
$s=S_2$ & 0 & 0.9 & 0.4 & indicative \\
$s=S_3$ & 0.4 & 0 & 0.9 & indicative \\
$s=S_4$ & 0.7 & 0.5 & 0.2 & direct \\
$s=S_5$ & 0.2 & 0.7 & 0.5 & direct \\
$s=S_6$ & 0.5 & 0.2 & 0.7 & direct \\
$s=S_7$ & 0.8 & 0.8 & 0.8 & indicative \\
\hline
\textbf{Regional type prior $P(t|r)$} & $t=A$ & $t=B$ & $t=C$ & - \\
\hline
$r=R_1$ & 0.6 & 0.3 & 0.1 & contextual \\
$r=R_2$ & 0.1 & 0.6 & 0.3 & contextual \\
$r=R_3$ & 0.3 & 0.1 & 0.6 & contextual \\
\hline
\end{tabular}
}
\vspace{-1.5mm}
\end{table}

\section{EVALUATION}\label{sec:evaluation}

\subsection{Baseline}

A simplification of the presented method was selected as a baseline for evaluation. It represents an intuitive ad-hoc method that relies on late fusion for classification, whereas the approach introduced in Sec. \ref{sec:type-classification} is based on early fusion of per-sensor threat type likelihoods. First, MAP estimation over all sensor-wise posteriors $P(t|Z^{(s)},r) \propto P(t|r) P(Z^{(s)}|t,r)$ is performed ($\forall s: Z^{(s)} \in \mathcal{Z}$):
\begin{equation}\label{eq:baseline-posterior}
    \hat{t}^{(s)} = \argmax_{t' \in \mathcal{T}}P(t=t'|Z^{(s)},r).
\end{equation}
The fused predicted type $\hat{t}$ is then determined by performing a majority vote over the sensor-wise type predictions $\hat{t}^{(s)}$, where the prediction with the maximum posterior is selected in case of a tie.

\subsection{Scenarios}\label{sec:eval-scenarios}

For evaluation, two simulation scenarios are defined: (i) a basic scenario, constructed specifically for an ablation study of the methodology, and (ii) a CBRNE scenario, based on sensor configurations and region types from practical experience. In the basic scenario, there are three different threat object types $\mathcal{T}_{\text{basic}} = \{A, B, C\}$, three region types $\mathcal{R}_{\text{basic}} = \{R_1, R_2, R_3\}$ and seven sensors $\mathcal{S}_{\text{basic}} = \{S_1,\dots,S_7\}$. In the CBRNE scenario, there are four different threat object types $\mathcal{T}_{\text{CBRNE}} = \{A, B, C, D\}$ and six sensors $\mathcal{S}_{\text{CBRNE}} = \{S_1,\dots, S_6\}$. The region types $\mathcal{R}_{\text{CBRNE}} = \{$\texttt{grassland}, \texttt{road}, \texttt{road junction}, \texttt{road bend}, \texttt{road overpass}, \texttt{roadside marker}$\}$ were identified as informative for the CBRNE scenario. The ROI of the CBRNE scenario is labeled as described in Sec. \ref{sec:contextual-evidence}, with the labeling shown in Fig. \ref{fig:osint-regions}. This information is in turn used to assign the appropriate region prior to the respective detections. The parameters for both scenarios are specified in Tab. \ref{tab:basic_params} and Tab. \ref{tab:cbrne_params}. All priors are set based on domain knowledge, but could also be derived empirically if appropriate datasets are available.

\begin{table}[h]
\caption{CBRNE scenario parameters.}
\label{tab:cbrne_params}
\centering
\resizebox{\columnwidth}{!}{%
\begin{tabular}{l|cccc|l}
\hline
\textbf{Sensor prior $P_D^{(s,t)}$} & $t=A$ & $t=B$ & $t=C$ & $t=D$ & Evidence level \\
\hline
$s=S_1$ & 0.9 & 0.2 & 0.6 & 0.6 & indicative \\
$s=S_2$ & 0 & 0 & 0.6 & 0 & indicative \\
$s=S_3$ & 0 & 0 & 0.6 & 0.4 & indicative \\
$s=S_4$ & 0.3 & 0.7 & 0.5 & 0.5 & direct \\
$s=S_5$ & 0.4 & 0.6 & 0.5 & 0.5 & indicative \\
$s=S_6$ & 0.3 & 0.3 & 0.4 & 0.7 & indicative \\
\hline
\textbf{Regional type prior $P(t|r)$} & $t=A$ & $t=B$ & $t=C$ & $t=D$ & - \\
\hline
$r=\texttt{grassland}$ & 0.5 & 0.5 & 0 & 0 & contextual \\
$r=\texttt{road}$ & 0.4 & 0.4 & 0.1 & 0.1 & contextual \\
$r=\texttt{road junction}$ & 0.05 & 0.05 & 0.6 & 0.3 & contextual \\
$r=\texttt{road bend}$ & 0.3 & 0.3 & 0.1 & 0.3 & contextual \\
$r=\texttt{road overpass}$ & 0.05 & 0.05 & 0.6 & 0.3 & contextual \\
$r=\texttt{roadside marker}$ & 0.025 & 0.025 & 0.9 & 0.05 & contextual \\
\hline
\end{tabular}
}
\vspace{-1.5mm}
\end{table}

\subsection{OSINT Contextual Evidence}\label{sec:contextual-evidence}

\begin{figure}[h]
  \centering
  \includegraphics[width=\columnwidth]{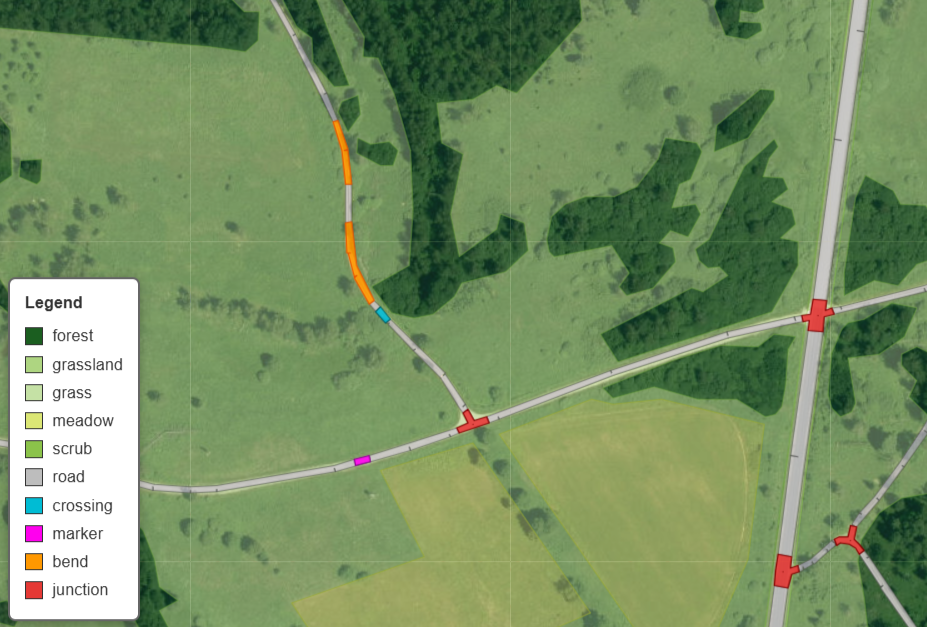}
  \caption{OSINT-derived regions for the CBRNE scenario.}
  \label{fig:osint-regions}
\end{figure}

The regions introduced in Sec. \ref{sec:problem-statement} as contextual evidence (Sec. \ref{sec:contextual-evidence}) are dependent on the use-case and can either be defined based on a priori information, or can be generated based on sensor data by performing e.g. scene segmentation, change detection or terrain analysis \cite{guo2020deep}. For the CBRNE scenario introduced in Sec. \ref{sec:eval-scenarios}, the region type labels are collected a priori from the GBD source OpenStreetMap (OSM)\footnote{https://openstreetmap.org/} and exploited as OSINT contextual evidence. By querying OSM through Overpass Turbo\footnote{https://overpass-turbo.eu/} and applying standard geographic information system (GIS) operations, the ROI can be labeled according to the region types $\mathcal{R}_{\text{CBRNE}}$. An in-depth description of the query structure and GIS operations involved in this process is beyond the scope of this work. The extracted regions are shown in Fig. \ref{fig:osint-regions} and the related regional priors $P(t|r)$ are shown in Tab. \ref{tab:cbrne_params}. Note that the region types \texttt{grass}, \texttt{grassland}, \texttt{meadow} and \texttt{scrub} have been merged into the region type \texttt{grassland}.

\begin{table*}[t]
  \centering
  \caption{Classification results for the simulation scenarios.}
  \label{tab:classification-results}
  \begin{filecontents*}{data/classification_results.csv}
,method,scenario,accuracy,precision,recall,f1,A,B,C,D
4,Proposed,Basic,0.95108,0.9510819450034442,0.9510795337450874,0.9510799635423758,0.9510935465037949,0.9508334584772489,0.9513128856460839,
1,Baseline,Basic,0.79331,0.7933076906207418,0.7933100288597221,0.7933082930158806,0.7927808293595703,0.7934116942427192,0.7937323554453523,
3,Proposed,Basic (w/o direct evidence),0.80711,0.8071136036551403,0.8071127515552748,0.8071103874512159,0.807950400072058,0.8070306993206257,0.8063500629609642,
0,Baseline,Basic (w/o direct evidence),0.63827,0.6382804373522429,0.6382686047386043,0.6382647667116709,0.6379095935154249,0.6369384459836721,0.6399462606359158,
5,Proposed,CBRNE,0.77293,0.8040953242462442,0.7734152315808559,0.770560987000421,0.8060528817948251,0.8330698568681906,0.7624035897354007,0.6807176196032673
2,Baseline,CBRNE,0.67591,0.7205414348052568,0.6757827556520364,0.6514324019539305,0.7841160165508376,0.8036892306399429,0.6633165829145728,0.3546077777103683
\end{filecontents*}
\pgfplotstabletypeset[
    col sep=comma,
    empty cells with={{n.a.}},
    columns={method, scenario, accuracy, f1, A, B, C, D},
    every head row/.style={
      before row=\toprule,
      after row=\midrule,
    },
    every nth row={2}{before row=\midrule},
    every last row/.style={after row=\bottomrule},
    every column/.style={
      fixed,
      fixed zerofill,
      precision=4,
      column type=c,
    },
    columns/method/.style={column name={\textbf{Method}}, string type, column type=l},
    columns/scenario/.style={column name={\textbf{Scenario}}, string type, column type=l},
    columns/accuracy/.style={column name={\textbf{Accuracy}}, column type=c},
    columns/precision/.style={column name={\textbf{Precision}}, column type=c},
    columns/recall/.style={column name={\textbf{Recall}}, column type=c},
    columns/f1/.style={column name={\textbf{F1}}, column type=c},
    columns/A/.style={column name={\textbf{F1 ($t=A$)}}, column type=c},
    columns/B/.style={column name={\textbf{F1 ($t=B$)}}, column type=c},
    columns/C/.style={column name={\textbf{F1 ($t=C$)}}, column type=c},
    columns/D/.style={column name={\textbf{F1 ($t=D$)}}, column type=c},
  ]{data/classification_results.csv}
\end{table*}

\begin{figure*}[t]
  \centering
  \begin{tikzpicture}
    \begin{groupplot}[
        group style={
          group size=3 by 1,
          horizontal sep=0.9cm,
        },
        bwplot,
      ]
 
      \nextgroupplot[
        title={(a) Sensor count},
        xlabel={Number of sensors $N_S$},
        ylabel={Classification accuracy},
        legend to name=leg1,
        legend columns=1,
        ymin=0.3, ymax=1,
      ]
      \addplot+[error bars/.cd, y dir=both, y fixed=0.015] table[col sep=comma, x index=0, y index=4] {data/sensor_sweep.csv};
        \addlegendentry{Proposed}
      \addplot+[error bars/.cd, y dir=both, y fixed=0.015] table[col sep=comma, x index=0, y index=3] {data/sensor_sweep.csv};
        \addlegendentry{Proposed (w/o dir. evidence)}
      \addplot+[error bars/.cd, y dir=both, y fixed=0.015] table[col sep=comma, x index=0, y index=2] {data/sensor_sweep.csv};
        \addlegendentry{Baseline}
      \addplot+[error bars/.cd, y dir=both, y fixed=0.015] table[col sep=comma, x index=0, y index=1] {data/sensor_sweep.csv};
        \addlegendentry{Baseline (w/o dir. evidence)}
 
      \nextgroupplot[
        title={(b) Clutter},
        xlabel={Clutter rate $\lambda$},
        xmode=log,
        log basis x=10,
        legend to name=leg2,
        legend columns=1,
        ymin=0.5, ymax=1,
      ]
      \addplot+[error bars/.cd, y dir=both, y fixed=0.015] table[col sep=comma, x index=0, y index=4] {data/clutter_sweep.csv};
        \addlegendentry{Proposed (strong conf. sep.)}
      \addplot+[error bars/.cd, y dir=both, y fixed=0.015] table[col sep=comma, x index=0, y index=8] {data/clutter_sweep.csv};
        \addlegendentry{Proposed (weak conf. sep.)}
      \addplot+[error bars/.cd, y dir=both, y fixed=0.015] table[col sep=comma, x index=0, y index=2] {data/clutter_sweep.csv};
        \addlegendentry{Baseline (strong conf. sep.)}
      \addplot+[error bars/.cd, y dir=both, y fixed=0.015] table[col sep=comma, x index=0, y index=6] {data/clutter_sweep.csv};
        \addlegendentry{Baseline (weak conf. sep.)}
 
      \nextgroupplot[
        title={(c) Prior perturbation},
        xlabel={Perturbation factor $\mu$},
        xmode=log,
        log basis x=10,
        legend to name=leg3,
        legend columns=1,
        ymin=0.5, ymax=1,
        cycle list={
            {black, solid,          mark=*,         mark size=1.2pt, mark options={solid}},
            {black, dashed,         mark=square*,   mark size=1.2pt, mark options={solid}},
            {black, dotted,         mark=triangle*, mark size=1.6pt, mark options={solid}},
            {black, dashdotted,     mark=diamond*,  mark size=1.6pt, mark options={solid}},
            {black, densely dashed, mark=otimes*,   mark size=1.6pt, mark options={solid}},
        },
      ]
      \addplot+[error bars/.cd, y dir=both, y fixed=0.015] table[col sep=comma, x index=0, y index=12] {data/prior_sweep.csv};
        \addlegendentry{Proposed (all priors pert.)}
      \addplot+[error bars/.cd, y dir=both, y fixed=0.015] table[col sep=comma, x index=0, y index=11] {data/prior_sweep.csv};
        \addlegendentry{Proposed (sensor priors pert.)}
      \addplot+[error bars/.cd, y dir=both, y fixed=0.015] table[col sep=comma, x index=0, y index=10] {data/prior_sweep.csv};
        \addlegendentry{Proposed (contextual priors pert.)}
      \addplot+[error bars/.cd, y dir=both, y fixed=0.015] table[col sep=comma, x index=0, y index=6] {data/prior_sweep.csv};
        \addlegendentry{Baseline (all priors pert.)} 
    \end{groupplot}
 
    \node[anchor=north] at ([yshift=-0.95cm]group c1r1.south) {\ref{leg1}};
    \node[anchor=north] at ([yshift=-0.95cm]group c2r1.south) {\ref{leg2}};
    \node[anchor=north] at ([yshift=-0.95cm]group c3r1.south) {\ref{leg3}};
  \end{tikzpicture}
 
 
  \caption{Ablation study with varying
           (a) sensor count,
           (b) clutter rate, and
           (c) prior perturbation.}
  \label{fig:sweep-results}
\end{figure*}
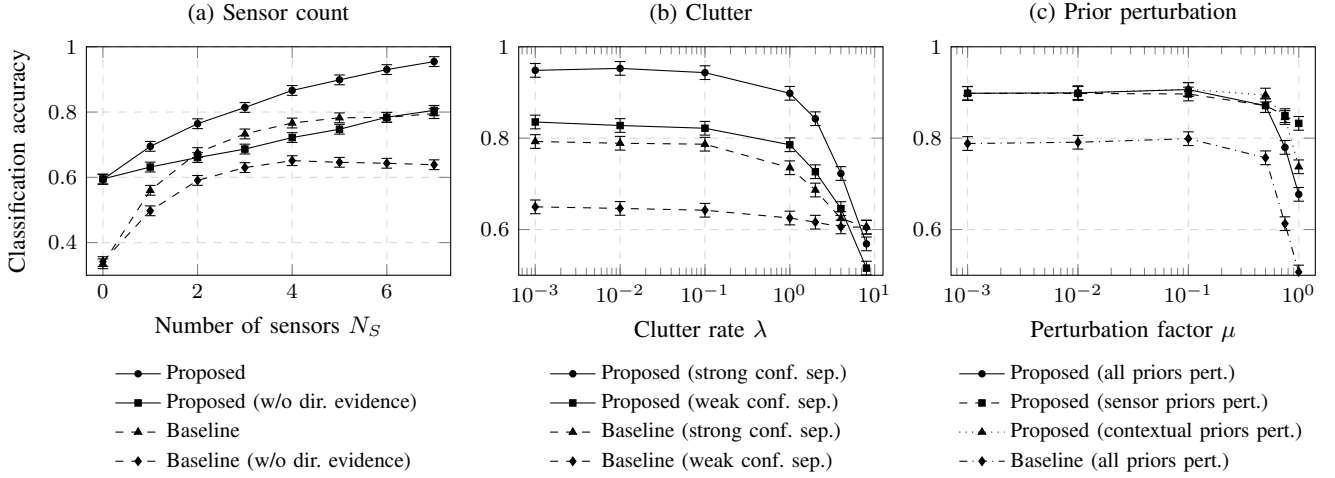

\subsection{Evaluation Method}\label{sec:eval-setup}

In this work, the evaluation of the Bayesian threat object classification method was performed via Monte Carlo simulations. The detection confidences of all sensors in both scenarios are drawn according to $\pi^{(s)} \sim \text{Beta}(8, 2.5)$. For each experiment, the respective metrics were computed over 10000 Monte Carlo runs. In each Monte Carlo run, a random threat object was created and a matching random detection set $\mathcal{Z}$ generated, based on the scenario and experiment specific parameters outlined in Sec. \ref{sec:eval-scenarios}. To evaluate the classifier, the actual object type was compared with the predicted type and the numbers of true and false predictions were recorded. For the simulations, a computation node with 2 AMD EPYC 9254 24-Core processors and 512 GiB RAM was used. 

For both scenarios, the classification performance was evaluated in terms of the macro-metrics accuracy and F1-score (balancing precision and recall), as well as in terms of per-class F1-scores. Additionally, an ablation study with the following experiments was carried out on the basic scenario:
\subsubsection{Sensor Number}

The number of available sensors $N_S \in \{0,\dots,|\mathcal{S}_{\text{basic}}|\}$ was varied, where for each Monte Carlo run $N_S$ were selected randomly from $\mathcal{S}_{\text{basic}}$, in order to study the influence of $N_S$ on the classification accuracy. Note that for $N_S=0$ there are no sensor detections, therefore the object type posterior is determined solely based on the regional prior. Furthermore, it was investigated how the performance degrades if all direct evidence sensors are replaced by indicative evidence sensors.

\subsubsection{Clutter Rate}

The influence of clutter detections on the classification accuracy was studied by varying the clutter rate $\lambda \in [0.001, 10]$. In each Monte Carlo run, $N_{\text{clutter}} \sim \min(\text{Poisson}(\lambda), |\mathcal{S}_{\text{basic}}|)$ sensor detections were assigned a clutter confidence $\pi^{(s)}_\text{clutter}$ (instead of a true detection confidence $\pi^{(s)}_\text{true}$) and in the case of direct sensor detections, a random predicted type. Furthermore, the influence of \textit{strong} ($\pi^{(s)}_\text{true} \sim \text{Beta}(8, 2.5),\ \pi^{(s)}_\text{clutter} \sim \text{Beta}(2.5, 8)$) and \textit{weak} ($\pi^{(s)}_\text{true} \sim \text{Beta}(5, 4),\ \pi^{(s)}_\text{clutter} \sim \text{Beta}(4, 5)$) confidence separation was investigated.

\subsubsection{Prior Perturbation}

In order to study how mismatches between true priors and priors used by the classifier affect the classification accuracy, for each Monte Carlo run, the classifier priors were perturbed with factor $\mu \in [0,1]$ \cite{huber1981robust}:
\begin{equation}
\begin{split}
        &[\Tilde{P}(t|R_1),\Tilde{P}(t|R_2),\Tilde{P}(t|R_3)]^T =\\
        &(1-\mu)[P(t|R_1),P(t|R_2),P(t|R_3)]^T+\mu\bm{\mathsf{u}},\\
\end{split}
\end{equation}
\begin{equation}
\begin{split}
        \Tilde{P}_D^{(s,t)} = \max(0, \min(1, P_D^{(s,t)}+\mu\mathsf{v})),
\end{split}
\end{equation}
with $\bm{\mathsf{u}}\sim \text{Dirichlet}(1,1,1)$ and $\mathsf{v} \sim \mathcal{U}(-1,1)$. Note that in the case of $\mu=0$ there is no perturbation, while the perturbation is strongest at $\mu=1$.

\section{RESULTS}\label{sec:results}

\subsection{Discussion}

The results shown in Tab. \ref{tab:classification-results} clearly illustrate that the proposed Bayesian object type classification method consistently outperforms the baseline in all scenarios and metrics, with a 15.8 percentage point increase in the basic scenario and a 9.7 percentage point increase in accuracy in the CBRNE scenario. Furthermore, it is demonstrated that the use of direct evidence strongly increases the classification performance compared to using only indicative evidence sensors. Overall, the performance is weaker in the more challenging CBRNE scenario, which is due to the fact that there is only one direct evidence sensor present and that there is one additional object type. The type-wise F1-score is uniform in the basic scenario, which is expected due to the symmetry of the sensor and regional priors. For the CBRNE scenario the type-wise F1-scores vary, with the classifiers performing best for type B and worst for type D, the reason being asymmetric sensor and regional priors.

Increasing the number of sensors used for fusion in the object type classification, as shown in Fig. \ref{fig:sweep-results} (a), steadily improves the classification accuracy of the proposed method, demonstrating the benefit of sensor fusion for the problem at hand. In contrast, the baseline method indicates a limited increase in classification accuracy for an increased number of sensors, showing the advantage of early fusion over late fusion. As noted above, having direct evidence sensors in addition to indicative sensors is beneficial for classification. In Fig. \ref{fig:sweep-results} (b), it is illustrated that both baseline and proposed methods remain robust to low and medium clutter rates, while performance degrades in high-clutter regimes. This requires strong clutter removal in the data association step prior to classification. Furthermore, a weak confidence separation between true and clutter detections has a strong negative impact, giving rise to the need for accurate confidence estimation on the sensor side. The effect of a mismatch between the priors used in the classifiers and the actual priors is shown in Fig. \ref{fig:sweep-results} (c). Both baseline and proposed methods remain robust in regimes of low and medium perturbation, while the classification performance declines strongly in regimes of high prior perturbation of both prior types. Furthermore, it is demonstrated that the performance decline remains limited in high prior perturbation regimes if only the sensor (direct and indicative evidence) or the regional (contextual evidence) priors are perturbed, while the other priors remain close to the true values. Additionally, it can be observed that the proposed method remains more robust towards regional prior perturbation with medium $\mu$, but performance declines more rapidly for high $\mu$ compared to sensor prior perturbation, which is likely due to the configuration of the scenario. 

\subsection{Limitations}

Limitations of the proposed Bayesian object type classification method include the fact that the conditional independence assumption with regard to the sensor detections is very strong and might not hold in all situations. Experiments with real data are required to verify whether this assumption is valid in practice. Furthermore, it must be stated that the sensor models used in the classification method are simplifications and $P_D^{(s,t)}$ would need calibration on physical data. Additionally, it should be emphasized that simulations cannot establish meaningful real-world performance but are a helpful tool for studying various performance characteristics of the proposed method.

\section{CONCLUSION}\label{sec:conclusion}

This paper proposed a context-aware, domain knowledge-enhanced Bayesian framework for heterogeneous multi-sensor CBRNE threat type classification. An evidence hierarchy formally separating direct, indicative, and contextual evidence was introduced, enabling integration of qualitatively different information sources within a unified probabilistic model. Contextual evidence was obtained from OSM and exploited as OSINT to introduce domain knowledge-informed regional type priors to the fusion process. A Bayesian MAP classifier that jointly exploits all evidence hierarchy levels was derived, with separate likelihood formulations for direct and indicative sensors.
Monte Carlo simulations of a basic scenario and a CBRNE scenario demonstrated that the proposed method consistently outperforms a majority-vote late fusion baseline, achieving a classification accuracy increased by up to 15.8 percentage points. The benefit of early over late fusion, the added value of direct evidence sensors, and robustness to moderate clutter rates and prior mismatch were confirmed in an ablation study. The method's computational complexity scales linearly in the number of threat objects, making it suitable for operational deployment.
Key limitations such as the conditional independence assumption with regard to sensor detections, the use of simplified sensor models, and the simulation-only evaluation still persist.

\subsection{Future Work}
Validation of the proposed evidence hierarchy with real-world CBRNE sensor data is the most critical next step, yet remains challenging due to the scarcity and sensitivity of labeled datasets in this domain. A comparison against learning-based classifiers faces the same data availability constraint. Although early fusion has been shown to be beneficial, its performance in practice depends critically on the quality of the pre-processing steps. Clutter mitigation, static object detection, and data association \cite{nausner2026soda} must be performed jointly and online prior to classification. Such a coupled pipeline may reveal additional challenges not captured in the present simulation study. Finally, applying parameter optimization to the proposed model — jointly over sensor priors and regional priors — offers a principled path toward deriving optimal sensor configurations and fusion system requirements for specific CBRNE use-cases. 

\addtolength{\textheight}{-12cm}   



%
%
\section*{ACKNOWLEDGMENT}
The authors used Opus-4.6 (Anthropic) for grammar and language editing assistance.




\bibliographystyle{IEEEtran}
\bibliography{references}

\end{document}